\documentclass[10pt,twocolumn,letterpaper]{article}

\usepackage{wacv}              

\usepackage[T1]{fontenc}

\definecolor{wacvblue}{rgb}{0.21,0.49,0.74}
\usepackage[pagebackref,breaklinks,colorlinks,allcolors=wacvblue]{hyperref}
\usepackage{enumitem}

\def\wacvPaperID{11} 
\def\confName{WACV}
\def\confYear{2026}

\title{SortWaste: A Densely Annotated Dataset for Object Detection in Industrial Waste Sorting}

\author{
Sara Inácio$^{1}$, Hugo Proença$^{1,2}$, João C. Neves$^{1,3}$\\[4pt]
\small $^{1}$University of Beira Interior, Portugal \quad \small $^{2}$IT: Instituto de Telecomunicações \small \quad $^{3}$NOVA LINCS\\
\small \texttt{sara.inacio@ubi.pt, hugomcp@ubi.pt, jcneves@ubi.pt}
}

\begin{document}
\maketitle
\begin{abstract}
The increasing production of waste, driven by population growth, has created challenges in managing and recycling materials effectively. Manual waste sorting is a common practice; however, it remains inefficient for handling large-scale waste streams and presents health risks for workers. On the other hand, existing automated sorting approaches still struggle with the high variability, clutter, and visual complexity of real-world waste streams. The lack of real-world datasets for waste sorting is a major reason automated systems for this problem are underdeveloped. Accordingly, we introduce SortWaste, a densely annotated object detection dataset collected from a Material Recovery Facility. Additionally, we contribute to standardizing waste detection in sorting lines by proposing ClutterScore, an objective metric that gauges the scene's hardness level using a set of proxies that affect visual complexity (e.g., object count, class and size entropy, and spatial overlap). In addition to these contributions, we provide an extensive benchmark of state-of-the-art object detection models, detailing their results with respect to the hardness level assessed by the proposed metric. Despite achieving promising results (mAP of 59.7\% in the plastic-only detection task), performance significantly decreases in highly cluttered scenes. This highlights the need for novel and more challenging datasets on the topic. 
\vspace{-0.6cm}
\end{abstract}
    
\section{Introduction}
\label{sec:intro}
\vspace{-0.1cm}
\begin{figure}
    \centering
    \includegraphics[width=\linewidth]{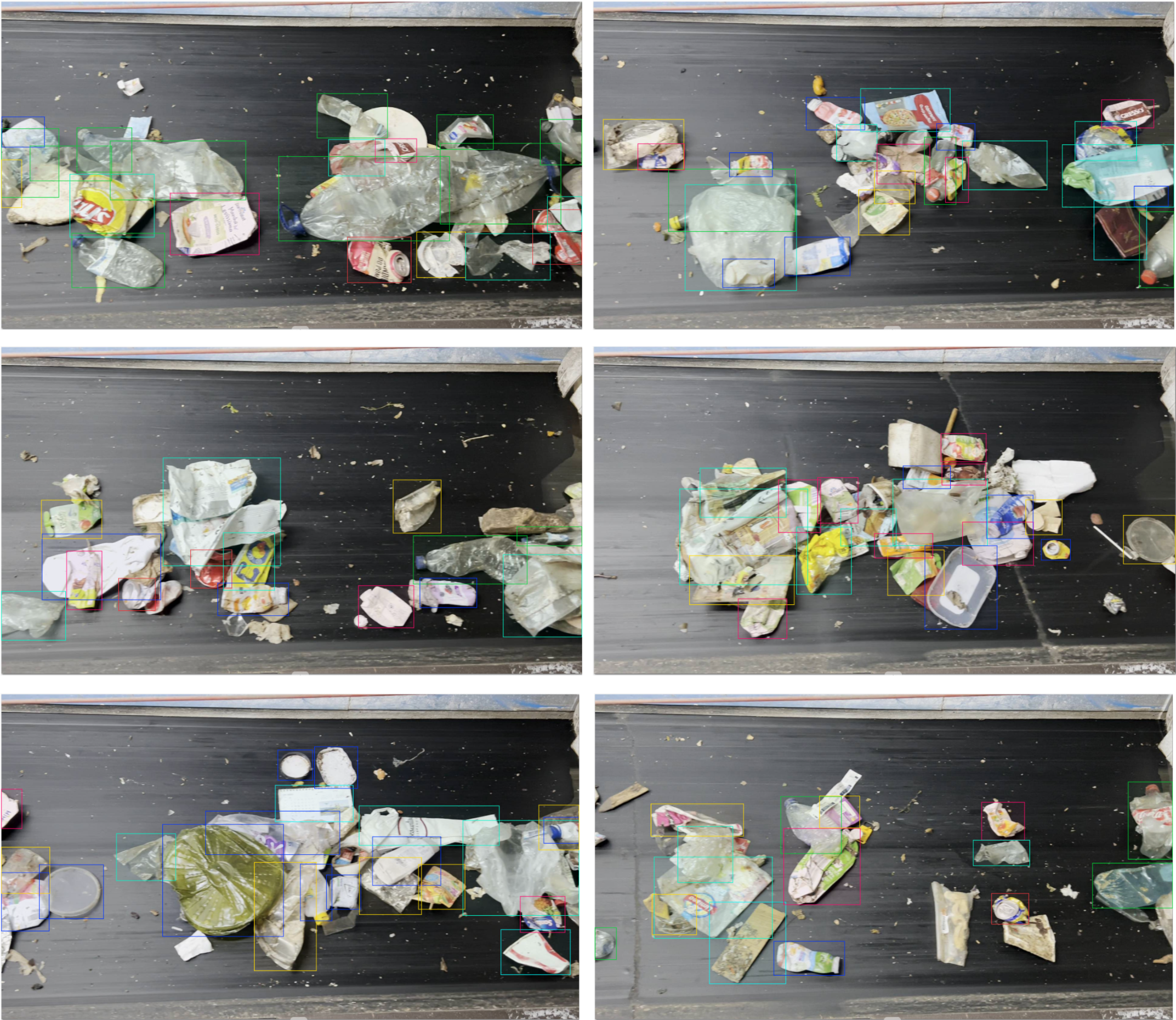}
    \caption{Sample images from the proposed waste dataset, illustrating diverse waste categories with bounding-box annotations for object detection.}
    \label{fig:enter-label}
\vspace{-0.3cm}
\end{figure}

The exponential growth of the global population and industrial development has increased waste production, making it one of the most pressing environmental challenges of the 21st century. Inefficient waste management systems negatively impact human health, ecosystems, water quality, and the environment at large \cite{b1}. As a result, developing innovative and more effective ways to manage and reduce waste is crucial to building a more sustainable future. 
Solid waste management is a key challenge, with global generation estimated at 2.01 billion tons annually and projected to reach 3.40 billion tons by 2050 under current trends \cite{b2}.

In most European countries, two principal systems are used for waste sorting. The first is a single-stream system, known as Mechanical-Biological Treatment (MBT), which processes Municipal Solid Waste (MSW). The second is a multi-stream system in which materials are selectively collected through source separation, such as plastics. The main challenge for the single-stream system is that valuable recyclables, particularly plastics, are often incorrectly discarded, limiting their recovery through proper sorting. This loss increases landfill use and consumes raw resources, sabotaging sustainability goals. However, as noted in \cite{b3}, MBT systems in urban areas can sometimes achieve recovery rates comparable to selective collection. This suggests that with improvements, MBT sorting could be highly impactful. 
Building on these challenges, deep learning and computer vision offer new opportunities for waste detection and recycling, improving efficiency, reducing reliance on manual labor, and mitigating health risks for workers exposed to hazardous materials. Nevertheless, automatic waste sorting remains challenging because real-world waste streams often contain objects that are deformed, broken, occluded, overlapping, or contaminated \cite{b13}.

Among the reasons for the lack of automation in waste sorting is the scarcity of publicly available datasets. Existing public datasets typically do not reflect the complex environments found in real-world waste sorting facilities \cite{b4, b5, b6, b7, b8, b9, b10, b11, b12}, except one contribution \cite{b13}. While some companies have developed proprietary datasets for internal use \cite{b26, b27, b28}, these datasets are not publicly released, restricting the scientific community's ability to advance research in this area.

This work introduces SortWaste, a densely annotated, publicly available dataset for object detection in waste sorting. Collected at an MBT facility, it captures real-world scenes with clutter, occlusion, deformation, and contamination — conditions especially challenging for computer vision systems. In addition to the scarcity of public datasets, there is no standard method to quantify the visual complexity of waste scenes, hindering objective dataset comparison and analysis of conditions that degrade model performance. To address this, we propose ClutterScore, a metric that measures visual clutter in a frame based on object count, class, and size entropy, and spatial overlap.

The main contributions of this work are:
\begin{itemize}[left=13pt]
    \item \textbf{SortWaste Dataset:} A publicly available, densely annotated dataset for object detection in MSW collected from an MBT facility. The dataset reflects real-world complexities, including cluttered scenes, overlapping and deformed objects, and dirty surfaces.
    \item \textbf{ClutterScore Metric:} A novel metric designed to quantify the visual complexity of waste scenes. We demonstrate its utility in analyzing the impact of clutter on object detection performance, showing that scene complexity is a critical factor in limiting model effectiveness.
    \item \textbf{Benchmarking with SOTA Models:} We evaluate several state-of-the-art object detection models, including Faster R-CNN, TridentNet, RetinaNet, and YOLOv11, on our dataset.
\end{itemize}
\noindent The project repository is available at \url{https://github.com/sarainacio/SortWaste}.
\vspace{-0.2cm}

\section{Related Work}
\label{sec:related_work}

\subsection{Waste Detection Datasets}
\vspace{-0.2cm}
Several public datasets exist for waste detection, classification, and recycling, covering general litter, marine debris, and industrial recycling. For clarity, we group them into four categories: Litter Detection, Marine Debris, General Waste Objects, and Recycling Waste. 

\vspace{0.35em}
\noindent\textbf{Litter Detection.} The Garbage in Photos (GINI) dataset \cite{b5} contains images of a single class. It was created using the Bing Image Search API and comprises approximately 1,400 images. Similarly, the WADE-AI dataset \cite{b6} includes pictures collected from Google Street View, providing a diverse visual set, though it is also limited to a single generic waste classification. Another relevant dataset is TACO (Trash Annotations in Context) \cite{b7}, introduced by Proença \textit{et al.}, designed for waste detection and segmentation. It contains 1,500 mobile-acquired images annotated into 28 waste categories and covers diverse real-world contexts, including outdoor environments, making it valuable to unconstrained applications. OpenLitterMap \cite{b9} is one of the largest public litter datasets, with over 100,000 user-contributed mobile images from around the world, providing diverse scenes and photographic styles that enhance its suitability for real-world litter detection.

\vspace{0.45em}
\noindent\textbf{Marine Debris.} The J-EDI dataset \cite{b14} consists of 5,720 images grouped into three classes, created to train deep learning models that enable autonomous underwater vehicles (AUVs) to detect waste in aquatic environments. In addition, AquaTrash, presented in \cite{b15} and derived from the TACO dataset, contains 369 multi-class annotated images. This dataset was developed to support AquaVision, a deep learning model aimed at detecting and classifying pollutants in the ocean and along the seashore, contributing to water pollution monitoring and mitigation.

\vspace{0.45em}
\noindent\textbf{General Waste Objects.} The Waste Pictures dataset \cite{b8} was constructed using images collected through Google searches. It contains approximately 24,000 images, organized into 34 categories representing a wide variety of waste types. The Taiwan Recycled Waste Dataset (TRWD) \cite{b12} comprises 6,233 images featuring multiple objects and is specifically tailored to the unique waste characteristics found in Taiwan. The dataset is divided into six classes.

\begin{figure}
    \centering
    \includegraphics[width=1\columnwidth]{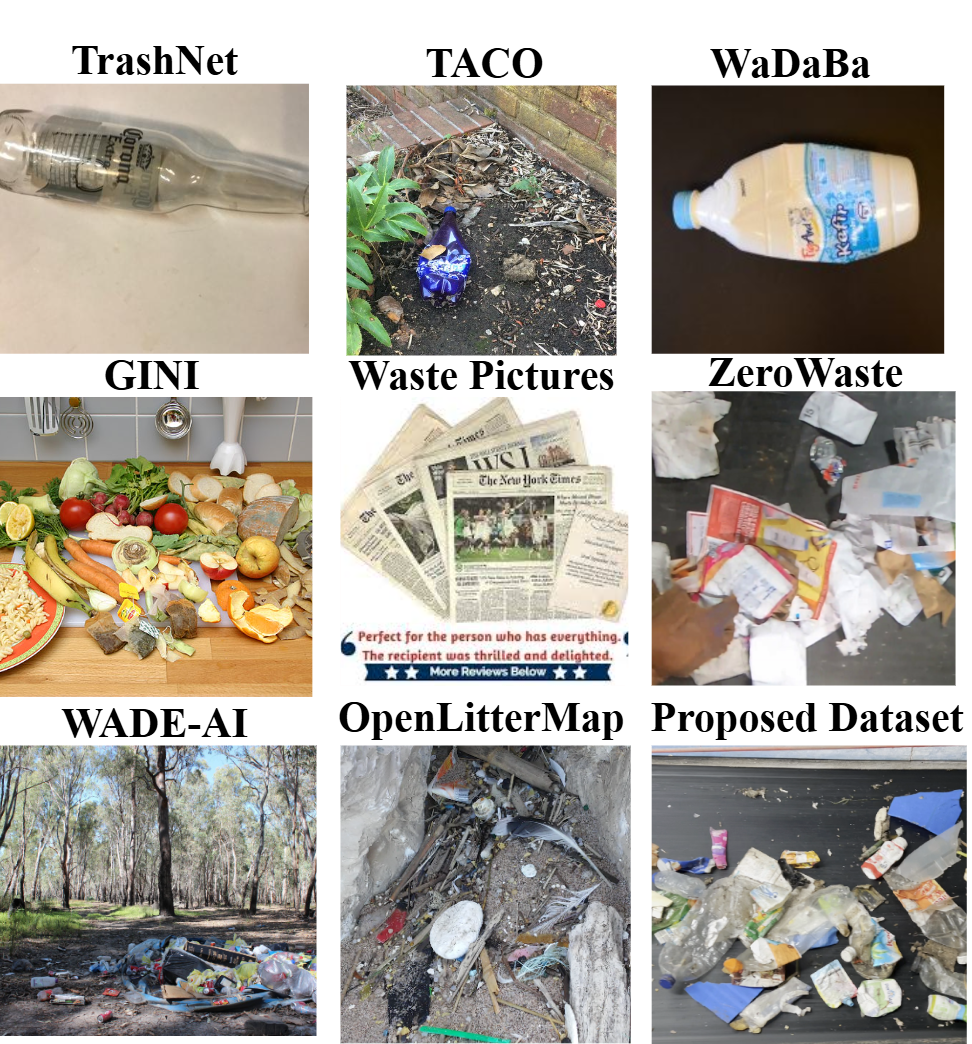}
    \caption{Representative samples from several state-of-the-art waste detection datasets. Each image illustrates a typical example from a different dataset, with the dataset titles displayed above the respective samples.}
    \label{fig:samples_sota_datasets}
\vspace{-0.5cm}
\end{figure}

\vspace{0.45em}
\noindent\textbf{Recycling Waste.}
Thung \textit{et al.} introduced TrashNet \cite{b4}, a widely used benchmark for waste classification comprising 2,400 images in six categories: metal, paper, plastic, glass, organic, and other, captured against a white background under varying lighting. The WaDaBa dataset \cite{b10} focuses exclusively on domestic plastic waste, with 100 plastic objects photographed 40 times each under different conditions, making it particularly relevant for controlled plastic waste identification. Labeled Waste in the Wild (LWW), introduced by Sousa \textit{et al.} in \cite{b11}, is a custom dataset containing 1,000 images of waste in food trays captured under real-world conditions. ZeroWaste-f \cite{b13} is one of the first high-quality waste datasets collected in real-world conditions. Acquired at a Material Recovery Facility (MRF) in the USA from a high-quality paper sorting line, it is split into three parts for different learning paradigms; the supervised ZeroWaste-f subset contains 4,503 annotated images with four classes.
\noindent While numerous datasets exist for waste detection tasks, most are not specifically designed for industrial applications, except for the ZeroWaste dataset \cite{b13}. Additionally, many datasets consist of images of a single object on a plain white background, which limits the generalizability and effectiveness of models trained on such data when applied to real-world environments. Table \ref{tab:summary-dataset} summarizes the reviewed datasets, and Figure \ref{fig:samples_sota_datasets} shows representative examples.

\begin{table*}[htp]
\centering
\caption{Comparison of public waste datasets.}
\label{tab:summary-dataset}
\begin{tabular}{lccccc}
\hline
\multicolumn{1}{c}{\textbf{Dataset}}               & \textbf{Images} & \textbf{Classes} & \textbf{Task}            & \textbf{Method}                                                                         & \textbf{Application}       \\ \hline
\multicolumn{6}{c}{\textbf{Litter Detection}}                                                                                                                                                                                             \\ \hline
\textbf{TACO \cite{b7}}           & 1500            & 28               & Segmentation             & Mask R-CNN                                                                              & Litter detection           \\
\textbf{WADE-AI \cite{b6}}      & 1396            & 1                & \begin{tabular}[c]{@{}c@{}} Detection +\\ Segmentation\end{tabular} & Mask R-CNN                                                                              & Litter detection           \\
\textbf{OpenLitterMap \cite{b9}}  & 100000+         & 100+             & Classification           & -                                                                                       & Litter detection           \\
\textbf{GINI\cite{b5}}           & 1400            & 1                & Classification           & -                                                                                       & Litter detection           \\ \hline
\multicolumn{6}{c}{\textbf{Marine Debris}}                                                                                                                                                                                                \\ \hline
\textbf{J-EDI \cite{b14}}         & 5720            & 3                & Detection                & \begin{tabular}[c]{@{}c@{}}YOLOv2, Tiny YOLO, \\ Faster R-CNN, SDD\end{tabular}         & Marine debris              \\
\textbf{AquaTrash \cite{b15}}     & 369             & 4                & Detection                & Faster R-CNN, RetinaNet                                                                 & Marine debris              \\ \hline
\multicolumn{6}{c}{\textbf{General Waste Objects}}                                                                                                                                                                                        \\ \hline
\textbf{TRWD \cite{b12}}          & 6233            & 6                & Detection                & YOLOv3                                                                                  & Waste detection            \\
\textbf{Waste Pictures \cite{b8}} & 23633           & 34               & Classification           & -                                                                                       & Trash objects              \\ \hline
\multicolumn{6}{c}{\textbf{Recycling Waste}}                                                                                                                                                                                              \\ \hline
\textbf{TrashNet \cite{b4}}       & 2400            & 6                & Classification           & SVM                                                                                     & Recycling waste            \\
\textbf{WaDaBa \cite{b10}}        & 4000            & 6                & Classification           & -                                                                                       & Recycling waste            \\
\textbf{LWW \cite{b11}}           & 1002            & 19               & Detection                & Faster R-CNN                                                                            & Recycling waste            \\
\textbf{ZeroWaste-f \cite{b13}}   & 4503            & 4                & \begin{tabular}[c]{@{}c@{}} Detection +\\ Segmentation\end{tabular} & \begin{tabular}[c]{@{}c@{}}RetinaNet, Mask R-CNN, \\ TridentNet\end{tabular}                                                        & Industrial recycling waste \\ \hline
\textbf{Proposed Dataset}                          & 5261            & 8                & Detection                & \begin{tabular}[c]{@{}c@{}}Faster R-CNN, RetinaNet, \\ TridentNet, YOLOv11\end{tabular} & Industrial recycling waste \\ \hline
\end{tabular}
\vspace{-0.2cm}
\end{table*}

\subsection{Object Detection Models}
Object detection is the task of identifying and localizing objects of interest within images or video frames. In this field, state-of-the-art approaches are categorized into two types: one-stage models and two-stage models.

\vspace{0.35em}
\noindent \textbf{Faster R-CNN} \cite{b17} is a two-stage object detector that extends R-CNN \cite{b16} and Fast R-CNN \cite{b18} by introducing a Region Proposal Network (RPN), which replaces hand-crafted methods with a fully convolutional network using anchor boxes across scales and aspect ratios. The generated proposals are refined by a Fast R-CNN \cite{b18} head for classification and bounding-box regression, and shared convolutional features between the RPN and the detector enable an efficient, end-to-end trainable pipeline that balances accuracy and speed.

\vspace{0.35em}
\noindent \textbf{Mask R-CNN} \cite{b18} extends Faster R-CNN \cite{b17} by adding a parallel branch for instance segmentation: for each Region of Interest (RoI), it predicts a binary mask alongside the class label and bounding box, enabling joint object detection and pixel-level instance segmentation.

\vspace{0.35em}
\noindent \textbf{Trident Network} \cite{b19} addresses the challenge of scale variation in object detection, which often limits the performance of models such as Faster R-CNN \cite{b17} and Mask R-CNN \cite{b18}. The key idea is to use trident blocks: a multi-branch architecture with shared parameters but different dilation rates. Each branch operates with a distinct receptive field, allowing the network to capture objects at different scales and perform scale-aware detection without relying on external image or feature pyramids. This design improves the performance of two-stage detectors while maintaining the inference complexity.

\vspace{0.35em}
\noindent \textbf{RetinaNet} \cite{b20}, introduced by Lin \textit{et al.}, is a one-stage detector that reaches accuracy comparable to state-of-the-art two-stage models. Its key contribution is Focal Loss, which mitigates the severe foreground–background imbalance in dense detection. The architecture combines a ResNet–FPN backbone with two subnetworks for anchor classification (using Focal Loss) and bounding box regression. This design substantially closes the accuracy and speed gap between one-stage and two-stage detectors.

\vspace{0.35em}
\noindent \textbf{YOLO} family of detectors \cite{b21} formulates object detection as a single-stage regression problem, enabling real-time performance. Successive versions of YOLO, from YOLOv1 to YOLOv10 \cite{b22,b23}, added features such as multi-scale detection, new activation functions, and anchor-free designs. The YOLOv11 \cite{b24} further improves efficiency and accuracy with components like the C3k2 block, SPPF module, and C2PSA mechanism, and extends the framework to instance segmentation, pose estimation, and multi-view recognition.

\section{SortWaste Dataset}
This section presents the SortWaste dataset, detailing data collection, annotation, preprocessing for dataset splits, and the proposed metric. 

\subsection{Data Collection Process}
\begin{figure}[h!]
    \centering
    \includegraphics[width=\linewidth]{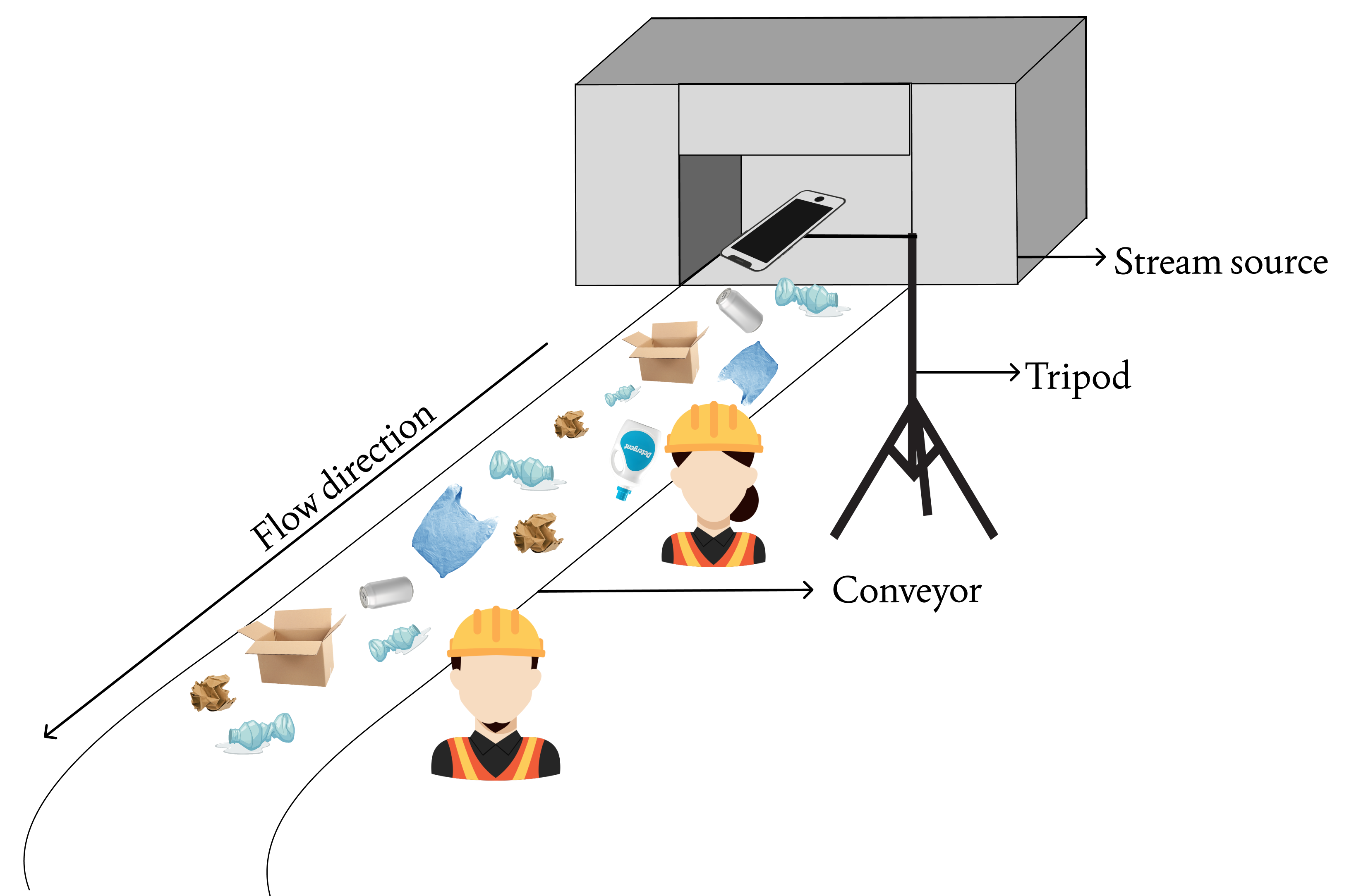}
    \caption{Illustration of the data collection setup, showing the arrangement of waste on the conveyor belt, the flow direction, the position of the workers, and the image capture system using a smartphone on a tripod.}
    \label{fig:data_collection}
\vspace{-0.4cm}
\end{figure}

The data in SortWaste were collected at an MBT facility. Manual sorting in this line aims to identify and remove items that can be recovered, such as different types of plastic, before the remaining waste is sent to landfills. Videos were collected using a smartphone mounted on a tripod beside the sorting line (Figure \ref{fig:data_collection}). The camera captured a top-down view of the conveyor belt from approximately 100 cm above the surface. The facility's ambient lighting was stable, so no additional illumination was required. All videos were recorded at 1920×1080 pixels and 60 fps. Data collection occurred at the beginning of the sorting line, before direct human intervention, but after several mechanical preprocessing steps applied between the arrival of the waste and the point of manual sorting. This setup allowed us to capture representative visual conditions of the operational environment while preserving normal workflow and complying with on-site safety and regulations.

\subsection{Annotation}
\begin{figure}[t]
    \centering
    \includegraphics[width=\linewidth]{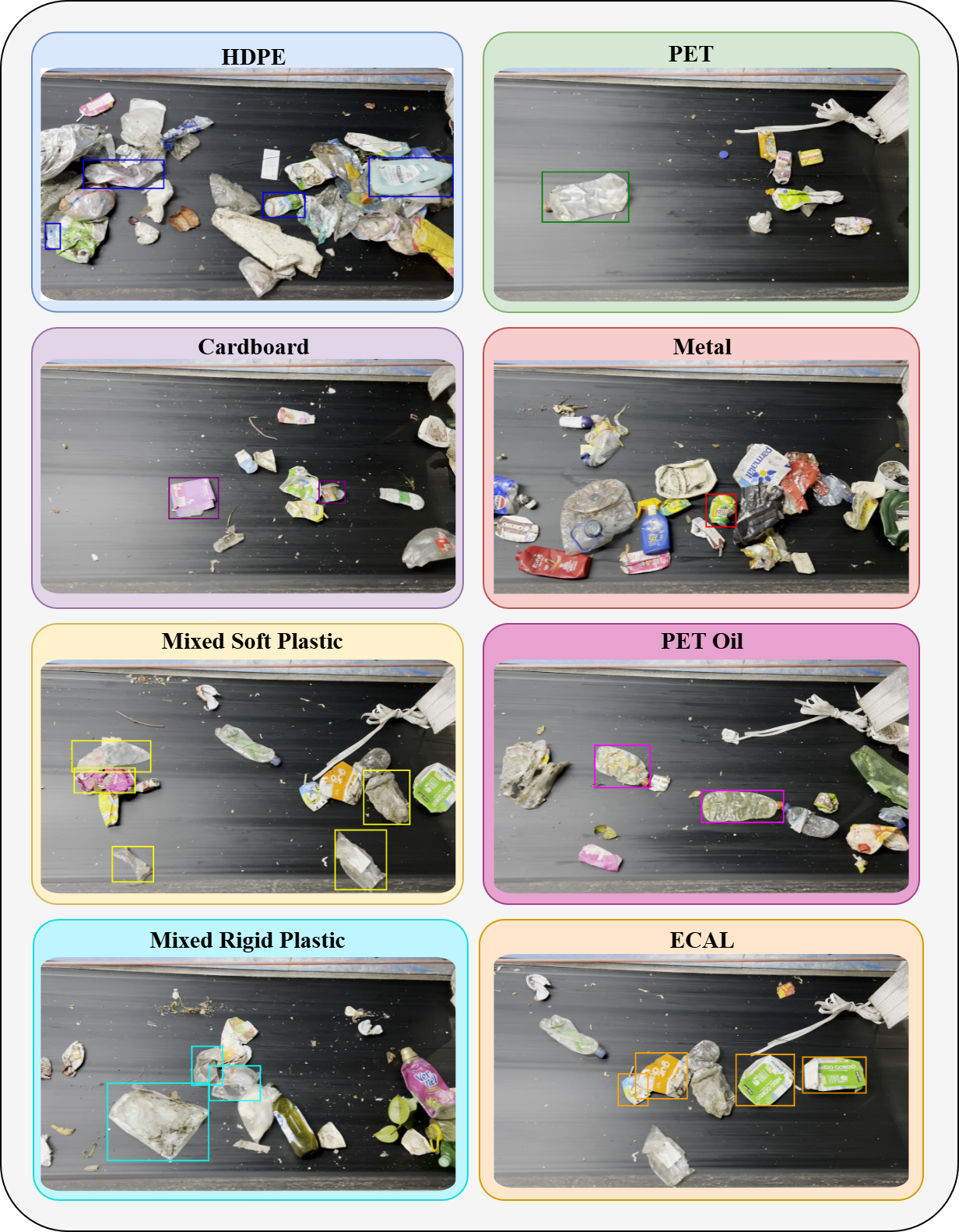}
    \caption{Representative examples from the SortWaste dataset, illustrating different categories of waste items on a conveyor belt.}
    \label{fig:dataset}
    \vspace{-0.4cm}
\end{figure}

The frame rate was downsampled to 5 fps to preserve variability while reducing annotation workload. In total, 18 minutes of video were annotated, yielding 5,396 frames with dirty, deformed, broken, and overlapping objects. Annotations were created over approximately two months using CVAT \cite{b29}, with each frame requiring, on average, six minutes, depending on scene complexity and object count. This careful annotation process was essential to ensure the quality of the densely labeled dataset. Although there are various types of plastic materials, some are challenging to distinguish using computer vision. Therefore, the annotation focused on simpler and more easily identifiable plastic classes.\\
Eight types of materials were annotated:
\begin{itemize}[left=13pt]
    \item \textbf{Polyethylene Terephthalate (PET):} Rigid, transparent or green objects, usually bottles, jars, and other containers previously used to package water, soft drinks, or
other beverages.
    \item \textbf{High-Density Polyethylene (HDPE):} Opaque, colored objects that are less flexible and denser, such as yogurt cups, bottles, and jars used for food products, hygiene items, detergents, fabric softeners, or alcohol.
    \item \textbf{Liquid Food Cartoon Packaging (ECAL):} Multilayer packaging composed of at least 75\% cardboard, intended for containing liquid foods (e.g., milk, juice).
    \item \textbf{PET Oil:} PET containers specifically used for packaging edible oils. Although this subcategory is not considered an independent category, it can be viewed as a subdivision of PET due to its typical contamination.
    \item \textbf{Mixed Soft Plastic:} Flexible and compressible plastics, such as cookie wrappers, potato chip bags, and plastic bags.
    \item \textbf{Mixed Rigid Plastic:} Rigid plastics that do not fall under the HDPE category, often transparent, such as molded packaging, boxes, and other hard containers.
    \item \textbf{Cardboard:} Corrugated or flat cardboard packaging used for storing, transporting, and distributing products.
    \item \textbf{Metal:} Metallic packaging made of steel or aluminum, such as beverage cans or food tins.
\end{itemize}
Figure \ref{fig:dataset} shows examples of all annotated classes in the SortWaste dataset.

\vspace{-0.1cm}
\subsection{Preprocessing}
To minimize overlap between subsets, the dataset was partitioned into scenes of 200 consecutive frames, and the first five frames of each scene were discarded to reduce temporal redundancy and ensure a more precise separation between adjacent scenes. Then, we grouped these scenes into approximately 70\%/15\%/15\% for training, validation, and test splits, respectively, while maintaining similar class distributions. Due to variation in object frequency, achieving target proportions exactly for each class was not possible, but the final splits closely match the desired percentages with comparable distributions, as shown in Table~\ref{tab:statisticsnew}.


\begin{table}[t]
\centering
\caption{Summary of object counts per class and per dataset split, including all object categories.}
\label{tab:statisticsnew}
\small
\begin{tabular}{c|ccc|c}
\hline
\textbf{Class Name}          & \textbf{Train} & \textbf{Validation} & \textbf{Test} & \textbf{Total} \\ \hline
\textbf{HDPE}                & 16803          & 4972                & 3269          & 25044          \\
\textbf{ECAL}                & 13649          & 2552                & 3026          & 19227          \\
\textbf{PET}                 & 11976          & 2108                & 2722          & 16806          \\
\textbf{Mixed Soft Plastic}  & 9077           & 1443                & 1817          & 12337          \\
\textbf{Mixed Rigid Plastic} & 7066           & 1120                & 1230          & 9416           \\
\textbf{Cardboard}           & 1524           & 425                 & 207           & 2156           \\
\textbf{Metal}               & 945            & 277                 & 215           & 1437           \\
\textbf{PET Oil}             & 802            & 168                 & 132           & 1102           \\ \hline
\textbf{\# Images}           & 3705           & 780                 & 776           & 5261           \\ \hline
\textbf{\# All Objects}      & 61842          & 13065               & 12618         & 87252          \\ \hline
\end{tabular}%
\end{table}

\begin{table}[t]
\centering
\caption{Summary of object counts per class and per dataset split for the plastic categories.}
\label{tab:statisticsnew_2}
\small
\begin{tabular}{c|ccc|c}
\hline
\textbf{Class Name}     & \textbf{Train} & \textbf{Validation} & \textbf{Test} & \textbf{Total} \\ \hline
\textbf{HDPE}           & 16803          & 4972                & 3269          & 25044          \\
\textbf{ECAL}           & 13649          & 2552                & 3026          & 19227          \\
\textbf{PET}            & 12778          & 2276                & 2854          & 17908          \\
\textbf{Mixed Plastic}  & 16143          & 2563                & 3047          & 21753          \\ \hline
\textbf{\# Images}      & 3705           & 780                 & 776           & 5261           \\ \hline
\textbf{\# All Objects} & 59373          & 12363               & 12196         & 83932          \\ \hline
\end{tabular}%
\vspace{-0.3cm}
\end{table}

\subsection{Statistics}

Table \ref{tab:statisticsnew} reports the number of annotated bounding boxes per class and per split for the full SortWaste dataset. The distribution is clearly imbalanced and reflects the operational context of the unsorted waste sorting line. Metallic objects are largely removed upstream using magnetic separators, which substantially reduces their occurrence in the recorded videos. Cardboard items also appear less frequently because they tend to disintegrate due to the stream's moisture. The PET Oil class is defined as a subcategory of PET and therefore occurs less often by construction. Overall, this imbalance simply reflects how often each material actually appears on the real sorting line. Table \ref{tab:statisticsnew_2} summarizes the statistics after regrouping the SortWaste into four plastic-centric categories. HDPE and ECAL are kept as separate categories. The PET category is expanded to include all instances originally labeled as PET Oil, given their similar material properties and visual appearance. The new Mixed Plastic category aggregates objects from the Mixed Soft Plastic and Mixed Rigid Plastic classes. Non-plastic materials are excluded from this analysis, as the focus is on plastic waste.

\begin{figure}[t]
    \centering
\includegraphics[width=1\linewidth]{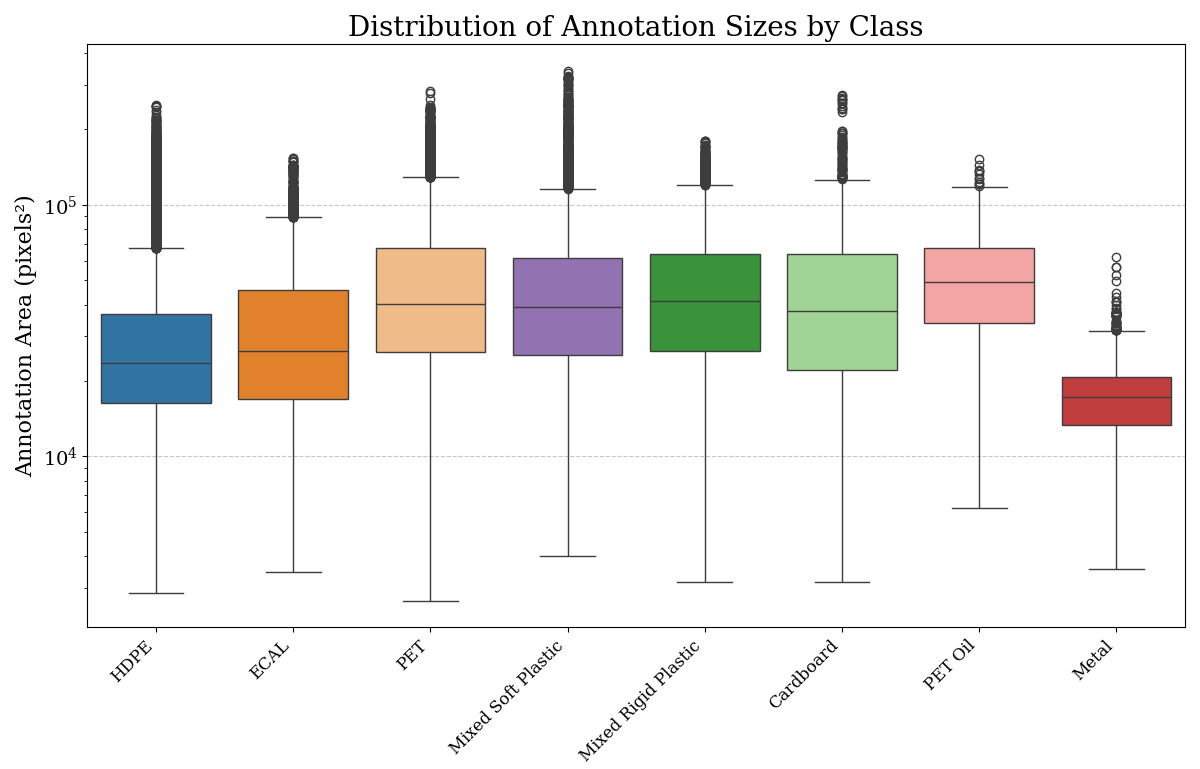}
    \caption{Distribution of annotation areas across waste classes. Boxplots show the variation in object sizes (in pixels²) for each category.}
    \label{fig:objects_area}
    \vspace{-0.4cm}
\end{figure}

Figure \ref{fig:objects_area} shows the distribution of annotation sizes across classes, with most exhibiting substantial size variability. PET, Mixed Rigid Plastic, and Cardboard have larger median areas, whereas Metal and HDPE are generally smaller. Numerous outliers, especially for PET and mixed plastics, reveal occasional very large instances, underscoring the heterogeneity of waste objects.

\begin{table}[t]
\centering
\caption{Comparison between the SortWaste and ZeroWaste-f datasets.}
\label{tab:summary-datasets}
\resizebox{\linewidth}{!}{%
\begin{tabular}{c|cc}
\hline
\textbf{Characteristic}       & \textbf{SortWaste}                         & \textbf{ZeroWaste-f}      \\ \hline
\textbf{Collection Location}  & Portugal                                   & Massachusetts, USA        \\
\textbf{Sorting Line Type}    & Municipal solid waste                      & High-quality paper stream \\
\textbf{Stream}               & Single                                     & Single                    \\
\textbf{\# Classes}    & 8                                          & 4                         \\
\textbf{\# Images}     & 5261                                       & 4503                      \\
\textbf{\# Bounding Boxes} & 87252                                      & 27744                     \\
\textbf{Material Diversity}   & \begin{tabular}[c]{@{}c@{}} High – various types\\ of plastics and waste\end{tabular} & Low – predominantly paper \\ \hline
\end{tabular}%
}
\vspace{-0.2cm}
\end{table}

Table \ref{tab:summary-datasets} compares SortWaste with ZeroWaste-f in terms of collection site, waste stream, number of classes and images, total bounding boxes, and material diversity, highlighting the main differences between our dataset and related work.

\subsection{ClutterScore: A Metric for Visual Complexity}
We propose ClutterScore, a metric for quantifying the visual complexity of each frame in our dataset, tailored for industrial waste-sorting scenarios. It provides a scalar measure of scene clutter, enabling fair comparisons across datasets and the analysis of model performance at different clutter levels. ClutterScore combines four terms: class distribution entropy, object count, object size entropy, and cumulative spatial overlap between bounding boxes, which together capture the main sources of visual clutter in waste-sorting scenes. Formally, we define:
\vspace{-0.1cm}
\begin{equation}
\textbf{ClutterScore} = \alpha \cdot H_c + \beta \cdot N + \gamma \cdot H_s + \delta \cdot O .
\label{eq:clutter}
\end{equation}
\vspace{-0.05cm}
The \textbf{class entropy}, \( H_c \), measures the uncertainty in object category distribution: \vspace{-0.2cm}
\[
H_c = \frac{-\sum_{i=1}^{C} p_i \log(p_i)}{\log(C)},
\]
where \( p_i \) denotes the proportion of objects belonging to class \( i \), and \( C \) is the total number of classes. Higher values of \( H_c \) correspond to frames with a more diverse mix of object types.

\noindent The \textbf{object count}, \( N \), captures how many objects are present in a frame, normalized across the dataset: \vspace{-0.2cm}
\[
N = \frac{N_o - N_{\min}}{N_{\max} - N_{\min}},
\]
where \( N_o \) is the number of objects in the current frame, and \( N_{\min} \), \( N_{\max} \) represent the minimum and maximum object counts observed across the dataset. This normalization ensures comparability between frames with varying density.

\noindent The \textbf{size entropy}, \( H_s \), accounts for the variability in object sizes:
\vspace{-0.1cm}
\[
H_s = \frac{-\sum_{d=1}^{10} q_d \log(q_d)}{\log(10)},
\]
where \( q_d \) is the proportion of objects falling into the \( d \)-th size bin, with the object size distribution discretized into 10 bins, greater entropy here implies a more diverse size range among objects.

\noindent Finally, the \textbf{bounding box overlap}, \( O \), captures the degree of spatial occlusion:
\vspace{-0.1cm}
\[
O = \frac{1}{A} \sum_{i \neq j} \text{area}(B_i \cap B_j),
\]
where \( B_i \cap B_j \) denotes the intersection area between the bounding boxes of objects \( i \) and \( j \), and \( A \) is the total area of the image in pixels. Higher overlap values indicate greater visual congestion in the scene.

All components are individually normalized to ensure that the final \textbf{ClutterScore} lies within the range $[0, 1]$. To balance the contribution of each factor to the overall measure of visual complexity, we derive its weights from the association value between each component and mAP@50. Specifically, we use the percentual importance of the Pearson correlations between ClutterScore and mAP@50. The obtained weights are $\alpha = 0.20$, $\beta = 0.32$, $\gamma = 0.15$, and $\delta = 0.33$, reflecting the relative influence of these four components on detection performance.

\subsection{Dataset Analysis with ClutterScore}

\begin{figure}[t]
    \centering
    \includegraphics[width=\linewidth]{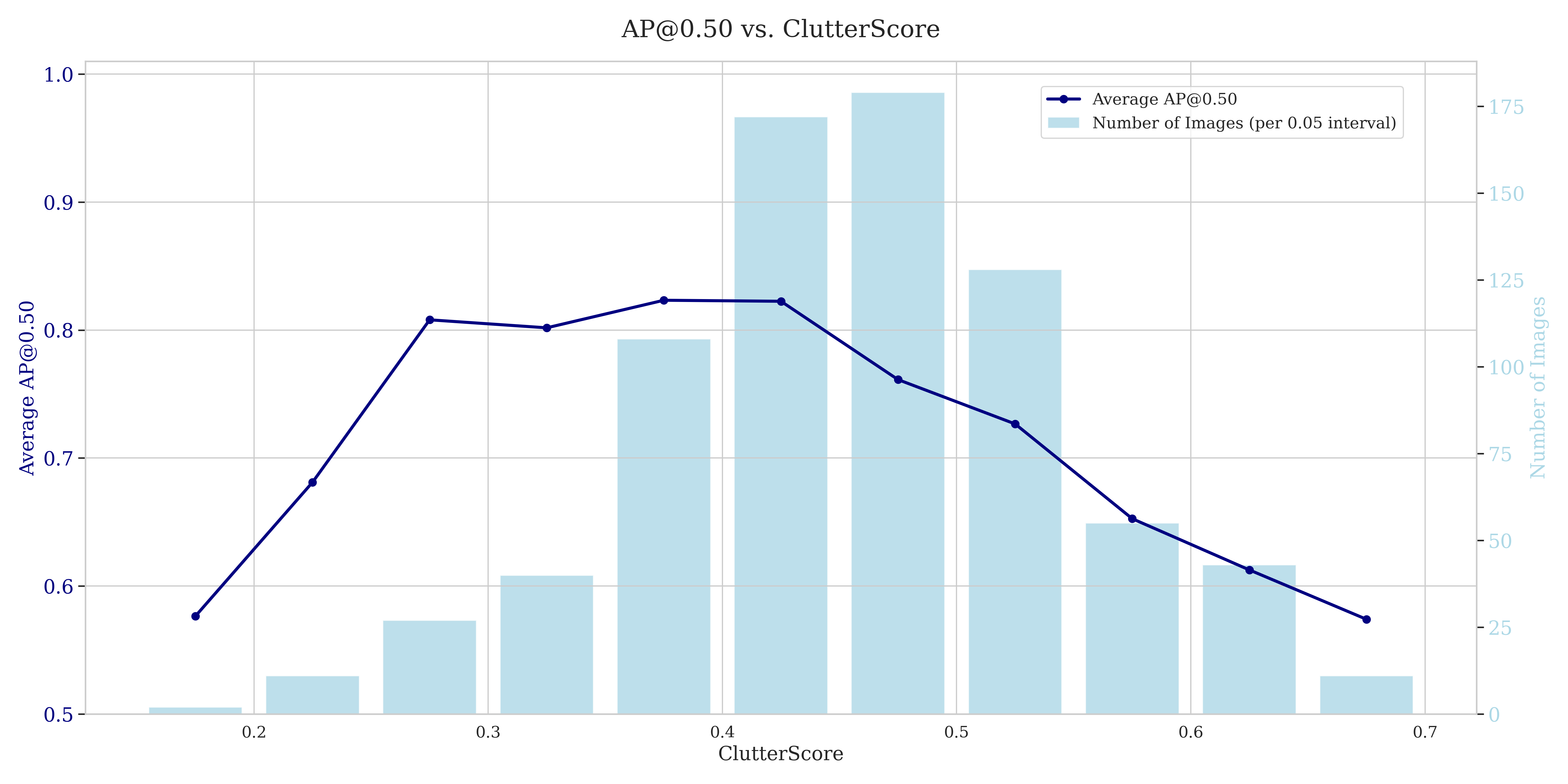}
    \caption{Relationship between ClutterScore and object detection performance. The line plot shows the average AP@0.50 for images grouped by ClutterScore, while the bars represent the number of images in each bin.}
    \label{fig:clutterscore_vs_map50}
\vspace{-0.1cm}
\end{figure}

Figure \ref{fig:clutterscore_vs_map50} shows the relationship between ClutterScore and model performance on the test set. Most frames fall within a medium complexity range (0.4–0.6). As ClutterScore rises above 0.4, detection performance degrades steadily, indicating that the visual complexity captured by the metric poses a significant challenge for object detection. However, clutter is not the only factor affecting performance. Uncontrolled object characteristics, such as broken, dirty, or deformed items, also increase detection difficulty and may explain low performance in some low-clutter images.

\begin{table}[t]
\caption{Average mAP@0.50 and ClutterScore components for high-clutter frames (ClutterScore 0.6–0.8).}
\vspace{-0.1cm}
\label{tab:components}
\resizebox{\linewidth}{!}{%
\begin{tabular}{c|c|c|c|c|c}
\hline
\textbf{Component}     & \textbf{mAP@0.50} & \textbf{\begin{tabular}[c]{@{}c@{}}Class\\ Entropy\end{tabular}} & \textbf{\begin{tabular}[c]{@{}c@{}}Object\\ Count\end{tabular}} & \textbf{\begin{tabular}[c]{@{}c@{}}Size\\ Entropy\end{tabular}} & \textbf{\begin{tabular}[c]{@{}c@{}}BoundingBox\\ Overlap\end{tabular}} \\ \hline
\textbf{Average Value} & 0.60              & 0.80                                                             & 0.91                                                            & 0.93                                                            & 0.14                                                                   \\ \hline
\end{tabular}%
}
\vspace{-0.3cm}
\end{table}

In the “High” clutter group, where performance drops to an mAP@0.50 of 0.60, visual complexity is mainly driven by high size entropy and object count (Table \ref{tab:components}). The normalized BoundingBoxOverlap term has a lower mean value (0.14) due to normalization rather than low importance: high object counts naturally lead to greater spatial overlap, making this term a direct consequence of object density.
\vspace{-0.2cm}
\section{Baseline Experiments}
\vspace{-0.1cm}
To evaluate the performance of state-of-the-art models on the SortWaste dataset, we design our experimental protocol in two distinct phases. In the first phase, referred to as \textbf{Full-Class Evaluation}, the models are trained and evaluated on the dataset using all annotated classes. In the second phase, termed \textbf{Plastic-Only Evaluation}, the models are trained and evaluated considering only the plastic-related classes.
\vspace{-0.5cm}

\subsection{Implementation Details}
\vspace{-0.1cm}
We use Faster R-CNN \cite{b17}, TridentNet \cite{b19}, and RetinaNet \cite{b20}, initialized with weights pretrained on the COCO dataset. For each model, only the learning rate and optimizer are tuned, while the batch size is fixed at eight. Training employs early stopping with a patience of 15 epochs or a maximum of 80,000 iterations. For YOLOv11 \cite{b24}, we adopt a similar tuning strategy, focusing on the learning rate and optimizer, with early stopping defined as 15 epochs of patience or a maximum of 300 epochs. These experiments aim to identify the optimal configuration and ensure fair comparisons across models.
\vspace{-0.1cm}

\subsection{Evaluation Metrics}
\vspace{-0.1cm}
We evaluate detection performance using mean Average Precision (mAP) as the primary metric. For each class, predictions are matched to ground-truth boxes based on the Intersection-over-Union (IoU), defined as the ratio between the area of intersection and the area of union of the predicted and ground-truth bounding boxes. A prediction is considered a true positive if its IoU exceeds a given threshold; otherwise, it is counted as a false positive. Average Precision (AP) is then computed per class from the corresponding precision–recall (PR) curve by aggregating precision values across recall levels. The mean Average Precision is obtained by averaging AP over all classes, using multiple IoU thresholds and confidence scores. In addition to the mAP values, we also report PR curves to visualize the trade-off between precision and recall and to provide a more detailed view of the detector’s behavior across confidence thresholds.

\subsection{Results}

\begin{table}[t]
\centering
\caption{Results on the SortWaste test set for state-of-the-art models fine-tuned on SortWaste. Best results are shown in bold.}
\vspace{-0.1cm}
\label{tab:results_all_classes}
\resizebox{\columnwidth}{!}{%
\begin{tabular}{c|cccc}
\hline
                             & \textbf{Faster R-CNN} & \textbf{TridentNet} & \textbf{RetinaNet} & \textbf{YOLOv11} \\ \hline
\textbf{PET}                 & 0.870                 & 0.854               & 0.844              & \textbf{0.880}   \\
\textbf{ECAL}                & 0.795                 & 0.778               & 0.785              & \textbf{0.808}   \\
\textbf{PET Oil}             & 0.652                 & 0.802               & 0.755              & \textbf{0.725}   \\
\textbf{HDPE}                & 0.700                 & 0.702               & 0.723              & \textbf{0.712}   \\
\textbf{Mixed Rigid Plastic} & 0.541                 & 0.547               & 0.562              & \textbf{0.568}   \\
\textbf{Mixed Soft Plastic}  & 0.460                 & 0.444               & 0.455              & \textbf{0.470}   \\
\textbf{Metal}               & 0.470                 & 0.419               & 0.517              & \textbf{0.330}   \\
\textbf{Cardboard}           & 0.093                 & 0.123               & 0.108              & \textbf{0.044}   \\ \hline \hline
\textbf{AP}                  & 0.415                 & 0.407               & 0.435              & \textbf{0.451}   \\
\textbf{AP50}                & 0.573                 & 0.584               & 0.594              & \textbf{0.567}   \\ \hline
\end{tabular}%
}
\end{table}

In Table \ref{tab:results_all_classes}, YOLOv11 \cite{b24} demonstrates the highest overall AP of 0.451, reflecting performance averaged across multiple IoU thresholds ranging from 0.5 to 0.95. Its strong performance on classes such as PET and ECAL, which possess highly distinguishable visual features, suggests its potential for reliable deployment in automated sorting systems. Contrarily, the low AP for Cardboard, likely due to its underrepresentation in the dataset, as shown in Table \ref{tab:statisticsnew}. Augmentation strategies may be critical for improving performance on low-frequency classes.

\begin{table}[t]
\centering
\caption{Results on the SortWaste test set for state-of-the-art models fine-tuned on SortWaste. Results are reported per plastic class. Best results are shown in bold.}
\vspace{-0.1cm}
\label{tab:results_plastic}
\resizebox{\columnwidth}{!}{%
\begin{tabular}{c|cccc}
\hline
                       & \textbf{Faster R-CNN} & \textbf{TridentNet} & \textbf{RetinaNet} & \textbf{YOLOv11} \\ \hline
\textbf{PET}           & 0.862                 & 0.836               & 0.858              & \textbf{0.872}   \\
\textbf{ECAL}          & 0.789                 & 0.774               & 0.775              & \textbf{0.786}   \\
\textbf{HDPE}          & 0.706                 & 0.714               & 0.692              & \textbf{0.729}   \\
\textbf{Mixed Plastic} & 0.623                 & 0.643               & 0.632              & \textbf{0.624}   \\ \hline \hline
\textbf{AP}            & 0.545                 & 0.519               & 0.550              & \textbf{0.597}   \\
\textbf{AP50}          & 0.745                 & 0.742               & 0.739              & \textbf{0.753}   \\ \hline
\end{tabular}%
}
\vspace{-0.4cm}
\end{table}

Table \ref{tab:results_plastic} shows results for plastic detection in MSW. The best detector achieves the highest overall AP of 0.597. This performance, coupled with a high AP50 of 0.753, demonstrates the model’s effective localization and high-confidence detections. The classes in this experiment were relatively balanced, as shown in Table \ref{tab:statisticsnew_2}, helping mitigate training bias. As in the previous experiment, PET remains the most detectable class due to its shape and brightness. In contrast, Mixed Plastic proved most difficult, mainly due to its visual heterogeneity and the merging of rigid and soft plastic types into a single class.

\begin{figure}[t]
    \centering
    \includegraphics[width=1\linewidth]{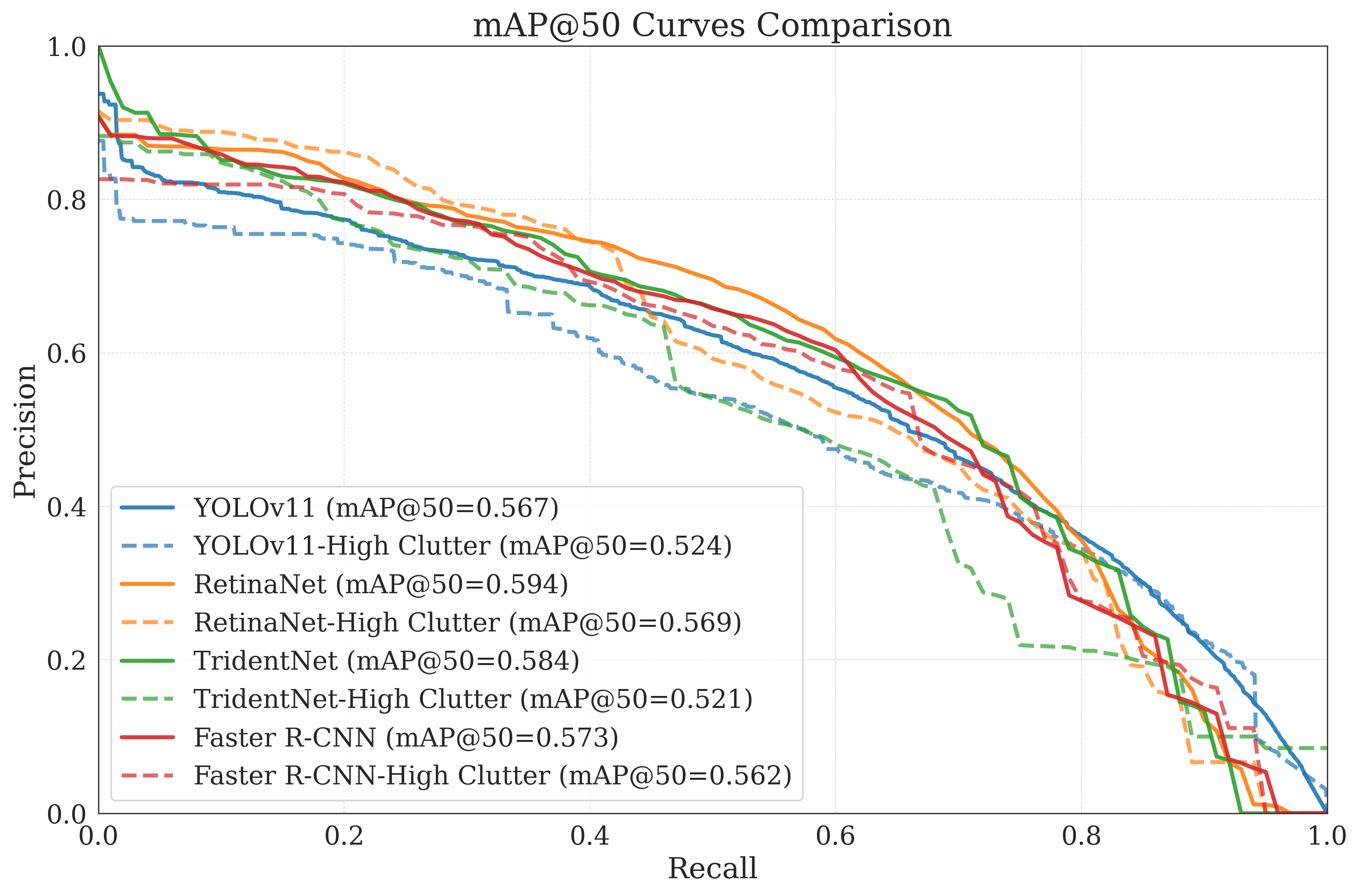}
    \caption{Precision-recall curves comparing mAP@0.50 across SOTA models on the full dataset (solid lines) and on high-clutter scenes (dashed lines).}
    \label{fig:pr_curves}
\vspace{-0.6cm}
\end{figure}

Figure \ref{fig:pr_curves} compares the behaviour of all evaluated models under varying scene complexity. Across architectures, we observe a decrease in precision and overall mAP@0.50 when moving from the full test set to the high-clutter subset, indicating that cluttered scenes remain challenging even for state-of-the-art detectors. This degradation suggests that current models are not robust to cluttered backgrounds and overlapping objects, and that improvements in object detectors should be targeted at these scenarios to ensure impact on real-world performance. While overall mAP scores provide a general benchmark, our focus is on assessing model robustness to visual clutter. To this end, we stratified the test set into different clutter-level subsets using our proposed ClutterScore metric. Qualitative examples illustrating the progressive increase in clutter across these subsets are provided in the supplementary material.
\vspace{-0.25cm}

\section{Impact and Limitations}
\vspace{-0.2em}
This work shows that object detection can improve waste sorting by increasing the quantity of recyclables, reducing worker exposure to hazardous materials, and improving productivity in Material Recovery Facilities (MRFs). The main contribution of this study is the introduction of the SortWaste dataset together with the ClutterScore metric. SortWaste addresses a gap in existing benchmarks by capturing the high density, occlusion, and heterogeneity of industrial waste streams. ClutterScore enables an analysis that goes beyond empirical evidence, quantitatively establishing that visual clutter is a key factor impairing detection. Figure \ref{fig:clutterscore_vs_map50} shows a performance drop associated with higher levels of clutter.

A limitation of SortWaste is class imbalance, which mirrors the natural material distribution in MSW. While this makes the benchmark more challenging, we argue that it is an essential property of a realistic dataset rather than a disadvantage. This characteristic opens up opportunities for future work on imbalance-aware learning, few-shot detection for underrepresented materials, and domain generalization under real industrial constraints. Although SortWaste is densely annotated, it contains fewer images than large-scale detection datasets, which may limit training high-capacity models from scratch. We therefore position it primarily as a realistic benchmark for fine-tuning, evaluation, and robustness analysis of pretrained models.
Finally, SortWaste was collected in a single MBT facility in Portugal, and waste appearance may vary across facilities due to collection schemes, local habits, processing configurations, and seasonality. As a result, the conclusions of this work may be limited to the conditions of the studied facility, being important that further studies build on our dataset to assess cross-site and cross-domain performance.
\vspace{-0.2cm}
\section{Conclusion}
\vspace{-0.2cm}
In this work, we introduced SortWaste, a densely annotated dataset for object detection in real-world waste-sorting environments. To address the key challenge of visual complexity in these scenes, we also proposed ClutterScore, a metric that quantifies scene clutter at the frame level. Our benchmarks with several state-of-the-art detectors show that automated plastic detection in industrial settings is feasible, with the best detector reaching a mAP of 59.7\%. On the other hand, our analysis shows that as scenes become more cluttered, model performance decreases, suggesting that current object detection models struggle at handling highly cluttered scenes. By making SortWaste publicly available and introducing ClutterScore, we aim to provide a foundation for future work and hope this will lead to research on models and training strategies that are not only accurate but also explicitly designed to be clutter-robust for practical waste management applications.
\vspace{-0.2cm}

\section*{Acknowledgments}
\vspace{-0.2cm}
This work is funded by national funds through FCT – Fundação para a Ciência e a Tecnologia, I.P., and, when eligible, co-funded by EU funds under project/support UID/50008/2025 – Instituto de Telecomunicações, with DOI identifier \url{https://doi.org/10.54499/UID/50008/2025}
This work is also financed by the project WATERMARK\footnote{WATERMARK project (Watermark-Based Algorithms for Trustworthy Media Authentication and Robust Certification in Public Administration), Project No. 2024.07356.IACDC, supported by “RE-C05-i08.M04 – Support the launch of a program of R\&D projects aimed at the development and implementation of advanced systems in cybersecurity, artificial intelligence, and data science in public administration, as well as a scientific training program,” under the Recovery and Resilience Plan (PRR), as part of the funding agreement signed between the Recovery Portugal Task Force (EMRP) and the Foundation for Science and Technology (FCT).} and supported by UID/04516/NOVA Laboratory for Computer Science and Informatics (NOVA LINCS) with the financial support of FCT.IP.



{
    \small
    \bibliographystyle{ieeenat_fullname}
    \bibliography{main}
}

\end{document}